# ARTIFICIAL INTELLIGENCE IN THE AUTOMATIC CODING OF INTERVIEWS ON LANDSCAPE QUALITY OBJECTIVES. COMPARISON AND CASE STUDY.

Mario Burgui-Burgui
*University of Alcalá*
*Madrid (Spain)*

## 1. INTRODUCTION

Artificial Intelligence (AI) is already revolutionising the way we work and conduct research, and its future impact is challenging to predict. Concerning qualitative content analysis, recent studies demonstrate its usefulness for coding research interviews, a fundamental tool for data collection across numerous academic disciplines (Lopezosa and Codina, 2023; Zhang *et al.*, 2023).

However, its use is incipient and there are still not many experiences in the scientific literature, despite the need to analyse and closely monitor the development of tools with the potential to bring about such profound changes. Consequently, this paper illustrates its practical application in a real case where interviews were initially manually coded using expert criteria.

These interviews were carried out as part of a broader study aimed at evaluating the changes in landscape quality that occurred on a small island in Cuba (Cayo Santa María) as a result of tourism development (Burgui *et al.*, 2018). The European Landscape Convention, a global reference on this subject, indicates that landscape planning and management must always be carried out with the participation of citizens (Council of Europe, 2000). This participation involves expressing the needs and expectations of the population with respect to their landscape environment, which should serve as support for experts and technicians to determine the Landscape Quality Objectives (LQOs) and subsequent planning and management measures. In this sense, the interviews included in this proposal were conducted to formulate the LQOs and specific measures for the landscapes of Cayo Santa María (Burgui *et al*., 2017).

The work presented here reviews the manual coding performed in the previous study, now comparing it with the automatic coding offered by three recently developed Artificial Intelligence (AI) tools. These tools include the AI service recently incorporated into the ATLAS.ti software and, on the other hand, Chat GPT and Google Bard. In all cases, the free version of these applications was examined, considering that their most widespread use in the short term for similar purposes will be that of students or novice researchers.

Atlas.ti is a CAQDAS software (Computer Aided Qualitative Data Analysis Software), specialised in qualitative content analysis. Widely used in the academic sphere since its launch in 1993, it introduced an AI functionality in March 2023 based on the GPT model by OpenAI. This feature facilitates the automatic establishment of categories and codes for the text under examination (Lopezosa *et al*., 2023).

Chat GPT (Chat Generative Pre-Trained Transformer) is an AI application in the form of a conversational bot or *chatbot*, categorized under "Large Language Models". Launched in 2022 by OpenAI, it specialises in dialogue, employing natural language processing (NLP) techniques to generate coherent, real-time responses in conversation format. At the time of this study (December 2023) it consists of the GPT-3.5 (free) and GPT-4 (subscription) models (OpenAI, 2022).

Google Bard, introduced by Google in March 2023 in response to ChatGPT's success, is a chatbot based on the PaLM 2 family, a language model for dialogue applications (Singh *et al*., 2023). On Google's website, Bard is defined as an AI experiment, a creative collaborator that helps increase productivity (Google, n.d.). As in the case of ChatGPT, the creators warn that these applications may give wrong answers or incorrect information.

The coding process is fundamental in qualitative research, as it involves the analysis of information to assign labels or codes to the most relevant aspects of the data, in order to highlight key themes, patterns or concepts (Skjott and Korsgaard, 2019). This is a process that usually requires large amounts of time, although there is software that has allowed some stages to be semi-automated (Chandra *et al.*, 2019). It is therefore not surprising that there is interest in further automating the process, and this is where AI can be useful.

However, as with any such advance, greater automation does not always lead to improved results, hence the importance of developing comparative studies between manual and automatic coding. In this regard, the scientific literature shows both supporters and detractors. The former argue that automatic coding is more reliable, objective and efficient (in terms of time) than manual coding, while the latter argue that this technology is still underdeveloped (Graaf and Vossen, 2013). Nevertheless, recent studies have found a high level of accuracy and consistency in AI coding compared to manual coding (Hacking *et al.*, 2023).

## 2. OBJECTIVES

With the development and increasing availability of language models through AI, the use of automatic coding in qualitative studies is anticipated to significantly rise. In this sense, the overarching goal of the presented work is to contribute to the critical analysis of AI's utility in automatically coding research interviews. This is achieved through the examination of a real case in which the interviews had previously been coded manually. Specific objectives include the following:

- Explore the possibilities of AI-driven automatic code selection with three widely used tools (Atlas.TI, ChatGPT and Google Bard).

- Conduct a comparative analysis to evaluate the advantages and disadvantages of each tool.

- Assess the usefulness of the aforementioned process and tools in a case study focused on landscape quality research.

## 3. METHODOLOGY

### 3.1. Manual coding

In order to properly contextualise the work presented here, it is convenient to begin by summarising the process followed in the preparation and development of the interviews, as well as the treatment of the information obtained without AI in the reference study. A summary diagram can be seen in Figure 1.

**FIGURE 1.** *General methodology for the preparation and execution of the interviews, and for the processing of the information obtained.*

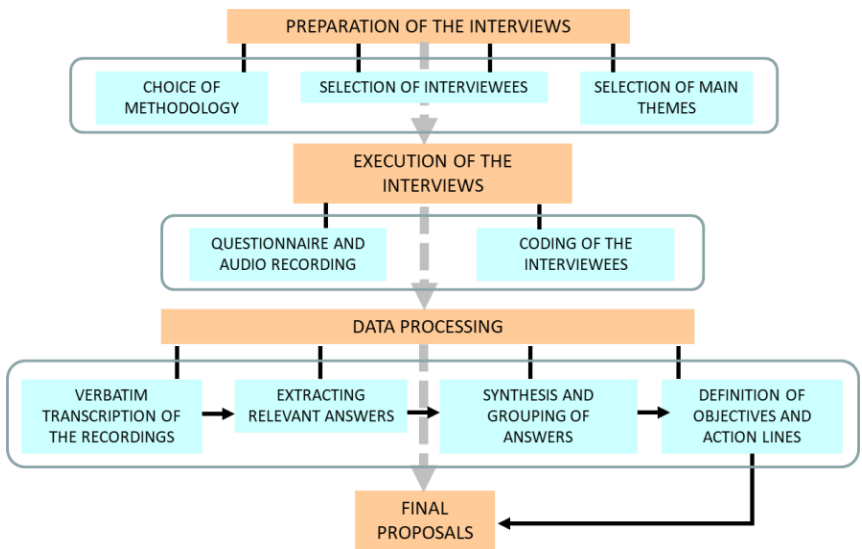

Source: the author.

The model chosen to obtain information was a semi-structured interview, with the following main characteristics:

– *Ad hoc* design (specifically for this study).
– Precise selection of objectives and interviewees.
– The interviewer determines the topic to be covered.
– There is an introduction explaining the purpose of the interview.

- The questions are pre-written and the respondent has to answer them within the framework of the question asked.
- There is an established order, but also a certain degree of freedom in the way the interview is conducted. The interviewee is also free to respond as he or she wishes (tone and extension).
- Within the established structure, the intention is to foster an open dialogue, with spontaneous, agile and dynamic conversations.
- Subsequent analysis will look at the content of both explicit and latent responses.

The selected interviewees were drawn from the primary sectors associated with the management and utilisation of the island: Tourism (National Tourist, International Tourist, Travel Agency Specialist, Manager of a hotel on the key); Environment (Professor of Biology at the University of Havana, Technician at the Cuban Environment Agency, Technician at the Center for Environmental Studies and Services of Villa Clara, Professor of Geography at the University of Havana, Researcher at the Institute of Tropical Geography); Others (Architect involved in the design of the hotels, Head of a department at the Institute of Physical Planning, Professor of Economics at the University of Havana). This approach aimed to ensure that the landscape quality objectives and proposed measures derived from the interviews were as comprehensive and varied as possible.

Based on what was proposed in the European Landscape Convention, the interview questions were structured around five distinct categories of Landscape Quality Objectives: 1) Conservation and maintenance of existing character, 2) Restoration of character, 3) Improvement of existing character, 4) Creation of new landscapes, 5) Promotion of landscape awareness and dissemination.

The chosen methodology for analysing and manually coding the interviews was Thematic Analysis, which consists of "classifying the text corpus by means of a set of representative themes related to the research objectives" (Conde, 2010). The theme is the central issue addressed in a text, which is normally developed through sub-themes.

As indicated, the themes corresponded to the 5 landscape quality objectives, while the subthemes were three: the reference of each objective to the main types of landscape units of the key (1: biotic and abiotic predominance, 2: anthropic predominance, 3: the entire key in general) (Figure 2).

**FIGURE 2.** *Map of Cayo Santa María with the general typology of landscape units: abiotic predominance, biotic predominance and anthropic predominance.*

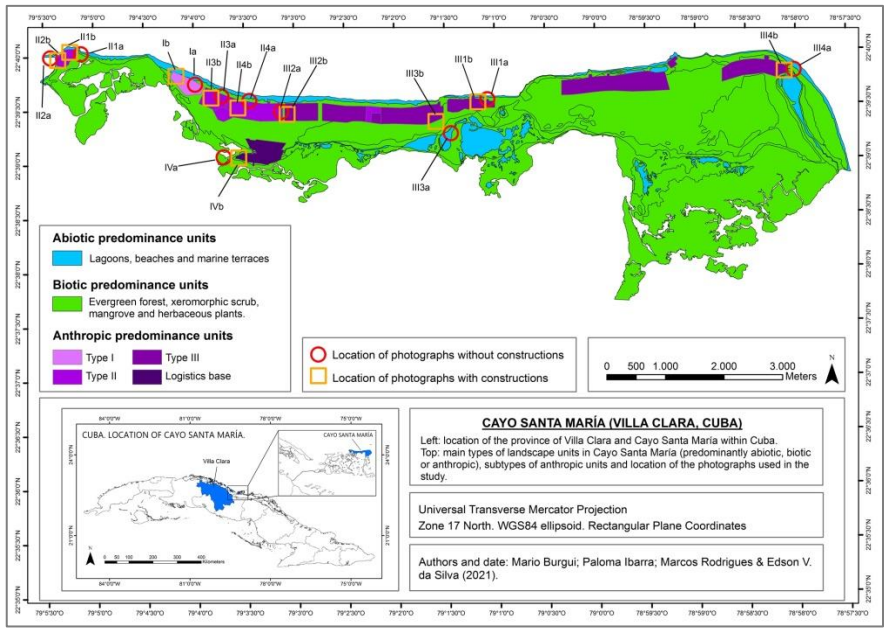

Source: The author.

Within the general framework of thematic analysis, the determination of topics and subtopics was also used, understood as fragments of text (such as sentences or paragraphs) addressing specific issues (Medina and Hernández, 2007). This approach is considered useful for extracting and categorising the information contained in a discourse (Soliveres *et al*., 2007). In summary, our methodology follows the illustrated structure in Figure 3. This scheme allowed a bidirectional transition between generality (Landscape Quality Objectives) and specificity (measures proposed by the interviewees for particular issues).

In this way it was possible to combine the deductive and inductive approaches. The former corresponds to the a priori identification of the 5 main issues (5 LQOs), while the latter entailed grouping specific measures into action lines and subsequently into LQOs. It is important to highlight that, although respondents were queried about LQOs in a general manner, their perspectives on the issues in the study area were often expressed through concrete examples (such as dirt, solid waste, lack of litter bins, etc.). This greatly enriched the corpus of information compiled, but required the inductive approach to group the very specific responses into the framework of general LQOs.

**FIGURE 3.** *Scheme for extracting and grouping information from the interviews.*

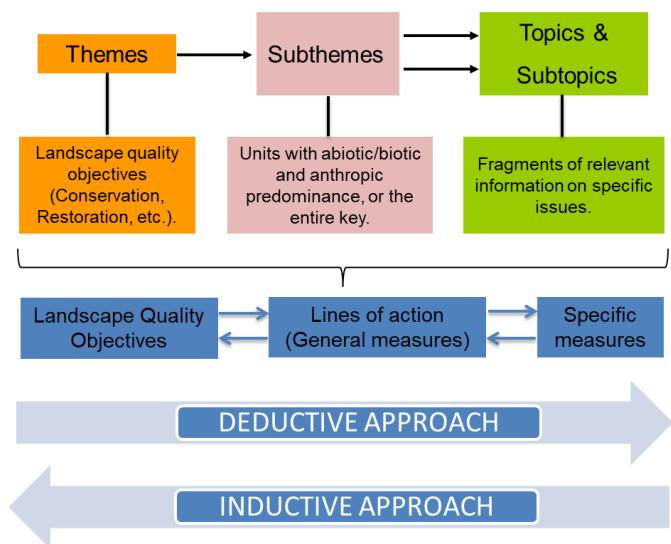

Source: The author.

On the basis of this scheme, the responses were coded and grouped to obtain a series of measures, lines of action and LQOs, which in turn were counted to determine the most numerous and those with the greatest consensus (Burgui *et al.*, 2017).

3.2. CODING USING ARTIFICIAL INTELLIGENCE

In order to conduct the automatic coding using AI, we first proceeded to remove from the transcript of the interviews both the questions and any other intervention of the interviewer, leaving only the interviewee's discourse. This text was incorporated into each examined tool and coding was subsequently requested, with some important differences that are worth noting. The first point to note is that, for this initial step of requesting automatic coding, Atlas.ti does not require a specific command that could affect the results, unlike ChatGPT and Google Bard. In Atlas.ti, after adding the text into the software, we simply navigate the menu for the options "Search and Code"➔ "AI Coding", and press the button to obtain categories and codes.

On the contrary, the outcomes from ChatGPT and Google Bard heavily rely on how the request is formulated, by means of an instruction or prompt. If this prompt is not properly formulated, the tool may return completely useless information. Therefore, it is advisable to follow guidelines for the proper wording of a prompt (White *et al.*, 2023) depending on the objectives pursued. In our case, after several trials, the following instruction was chosen:

> Imagine that you are a sociologist and you are conducting research on people's opinions and expectations about landscape. To do this, you have conducted twelve interviews that you need to analyse.
>
> From the transcript of the interviewees' responses, identify codes and subcodes for a qualitative analysis of the information.
>
> Please note that the questions asked by the interviewer are not included in the transcript.
>
> The transcript is as follows:

This prompt has a structure that includes the main segments of the instruction that the tool needs to properly process the information:

- Role: sociologist (must analyse this text from the perspective of a technician).

- Context: research on landscape views and expectations.

- Input: transcription of interviews without the intervention of the interviewer.

- Objective or specific request: codes and sub-codes for a qualitative analysis.

It should be noted that the components of the prompt can vary depending on the objectives pursued (White *et al.*, 2023). In our case, the instruction could have been even more specific, such as requesting ChatGPT and Google Bard to provide a specific number of codes and subcodes. However, this aspect was not included in the prompt mainly because the free version of Atlas.ti used in this study did not consistently allow the selection of the number of categories and codes (experiencing errors in this regard). Therefore, to ensure a fair comparison among the three tools, the most logical approach was to permit each tool to set the number of codes and subcodes freely. Consequently, the AIs were not directed to follow a specific response format (such as numbering or hierarchy).

Concerning the specific request, although Atlas.ti provides a classification into categories and codes, the instruction "identify codes and subcodes" was chosen. This decision stemmed from multiple tests with the transcripts of our study, revealing that ChatGPT and Google Bard appear to process the command more effectively when asked for codes and subcodes (corresponding to categories and codes in Atlas.ti, respectively).

3.3. COMPARISON CRITERIA

To evaluate the coding obtained with AI tools compared to manual coding, several criteria were defined based on a review of related literature (Hacking *et al.*, 2023; Moreiro, 2002), as well as others of own authorship applicable to this particular case:

1. Accuracy.

Definition: The ability of the AI to select the same codes as in manual coding.

Calculation: Percentage of codes and subcodes offered by AI that match those assigned manually.

2. Comprehensiveness.

Definition: The ability of AI to identify and code all relevant information in the interview.

Calculation: Percentage of correct codes and subcodes provided by AI with respect to the total number of Landscape Quality Objectives (LQOs) and related topics.

3. Thematic Coherence.

Definition: The logical relationship between codes and subcodes.

Calculation: Percentage of subcodes correctly assigned to their respective codes within each AI tool.

4. Redundancy.

Definition: Repetition or reiteration between codes and/or subcodes. Evaluated at all coding levels (code-code, code-subcode or subcode-subcode).

Calculation: Percentage of duplicated codes and subcodes in relation to the total number provided by the AI.

5. Clarity.

Definition: Simplicity, organization, and ease of interpretation of the codes and subcodes presented by AIs in their responses.

Weighting: Likert scale from 1 (Very Low) to 5 (Very High).

6. Detail.

Definition: The extent to which AIs provide explanatory details for codes and subcodes.

Weighting: Likert scale from 1 (Very Low) to 5 (Very High).

7. Regularity.

Definition: Criterion divided into two aspects. Firstly, by comparing the outcomes of the 12 interviews within the same test. Secondly, by comparing between different tests (distinct attempts with the same prompt) for the same interview.

Weighting: Likert scale from 1 (Very Low) to 5 (Very High).

Criteria 1 to 4 were calculated by percentage in each of the 12 interviews, and then an average of the overall percentage was obtained. Criteria 5 to 7 were weighted in a more qualitative way, using a Likert scale from 1 (Very Low) to 5 (Very High) in each of the interviews, to also obtain an overall average (e.g.: a value of 2 for criterion 4 would indicate that the AI response has "low" clarity).

All the aforementioned criteria were weighted from the point of view of the expert responsible for the manual coding, taking this as the correct reference. In this sense, it could be stated that a *cross-sectional expert criterion* was employed.

In contrast to other studies where criteria weighting was automated through computer programs (Hacking *et al*., 2023), with the expert criterion being an additional aspect, in this case the expert criterion served as a general weighting guideline for the rest of the criteria. This decision was made on the premise that a weighting approach based on in-depth knowledge of the subject is preferable to a mathematical weighting through specific software.

It was assumed that the manual coding conducted in the original study was accurate (even taking into account its limitations and potential inaccuracies), because it was based on a priori delimitation of themes and sub-themes (codes and subcodes: LQOs and landscape units, respectively), which in principle should also facilitate coding by AIs.

Finally, the advantages and disadvantages of the three tools were compiled in a summary table, based on the insights acquired from the aforementioned tests.

## 4. RESULTS AND DISCUSSION

### 4.1. WEIGHTING OF THE CRITERIA

Figure 4 shows a synthesis of the codes (Landscape Quality Objectives) obtained by manual coding in the baseline study. Below are the results of the weighting of the defined criteria, comparing the performance of automatic coding using AI[1] with that performed manually.

**FIGURE 4.** *Landscape Quality Objectives obtained by manual coding in the previous study.*

| CODE | TYPE | OBJECTIVES |
|------|------|------------|
| C1 | Conservation | 1. Landscapes that preserve natural values and elements, with adequate ecosystem and landscape functioning. |
| C2 | Conservation | 2. Landscapes that guarantee the sustainability and profitability of tourist activity. |
| C3 | Conservation | 3. Planning and management of tourist activity that gives priority to the conservation of landscapes. |
| I1 | Improvement | 4. Infrastructure and tourist constructions integrated into the landscape. |
| I2 | Improvement | 5. Landscapes free of pollution and impacts, both visual and those related to solid, liquid or gaseous waste. |
| I3 | Improvement | 6. Adequate management of natural resources and energy in the key. |
| I4 | Improvement | 7. A diversified tourist activity that respects the values and functioning of the landscape. |
| R1 | Restoration | 8. Comprehensive restoration plans adapted to the characteristics of the key, both for the pre-construction and construction stages, as well as for the management and dismantling stages. |
| S1 | Awareness-raising | 9. Ensure that planners, managers, technicians, workers, local people and tourists understand the importance (environmental, social and economic) of landscape conservation. |
| S2 | Awareness raising | 10. Achieve the dissemination of a tourism product in accordance with the landscape values of the area. |

Source: Burgui *et al*. (2017).

The *Accuracy* analysis clearly indicates lower results for Atlas.ti, with an average of 9.73% in the valid codes and 25.91% in the valid sub-codes of the total provided by the AI incorporated in this software (Figure 5). The two chatbots significantly outperformed Atlas.ti, demonstrating similar average values to each other, although the accuracy did not reach 50% in either case. This may seem a low percentage, but it is essential to consider the difficulty involved in the subject matter addressed in this research (the landscape quality objectives are not a common topic). For this reason, it is surprising that in some cases success values approached 70% (Interview 9) or even reached 100% (Interview 4). As an example, in Interview 4 ChatGPT included the following four codes in the answer: "Landscape Conservation", "Restoration and Maintenance", "Dissemination and Education Objectives", "Landscape Creation".

---

[1] It should be noted that the free version of Atlas.ti only permitted the analysis of 9 out of the 12 interviews during the test period. Nevertheless, this limitation did not hinder the evaluation of all the established criteria and the comparison of its responses with those of the other AIs in all the tests.

**FIGURE 5.** *Percentage of Accuracy of codes (top) and subcodes (bottom) assigned by the AIs in each interview. The legend shows the average percentage of Accuracy for the set of interviews.*

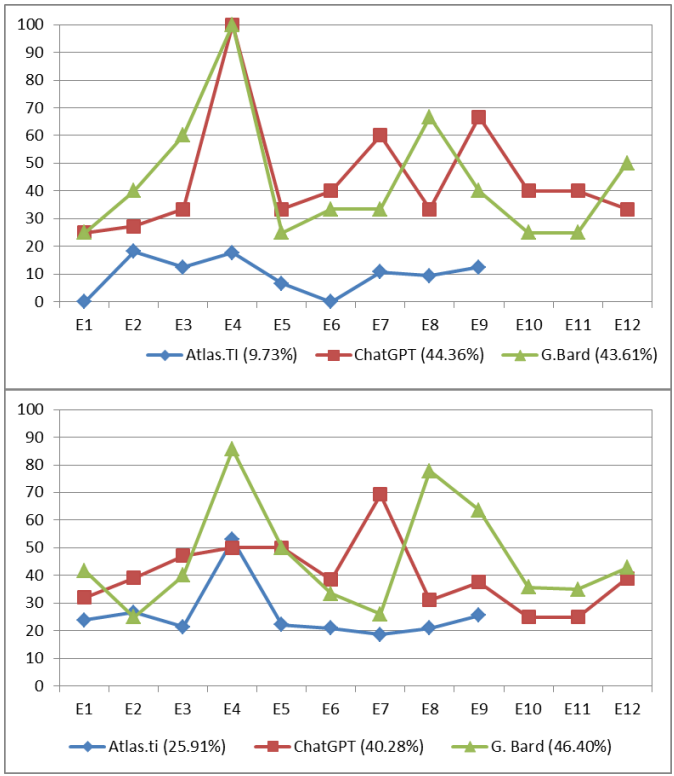

Source: The author.

*Comprehensiveness* is a criterion that complements the preceding one, as it may happen that an AI offers a result with a high percentage of precision in the codes or subcodes, relative to the overall response, but which nevertheless does not cover a sufficient percentage of the relevant information under study. This is what explains the low comprehensiveness values observed, for example, in certain Google Bard responses that hover around 20%.

**FIGURE 6.** *Average percentage of the criterion Comprehensiveness for codes (top) and subcodes (bottom) assigned by the AIs in each interview. The legend shows the average percentage of this criterion for the set of interviews.*

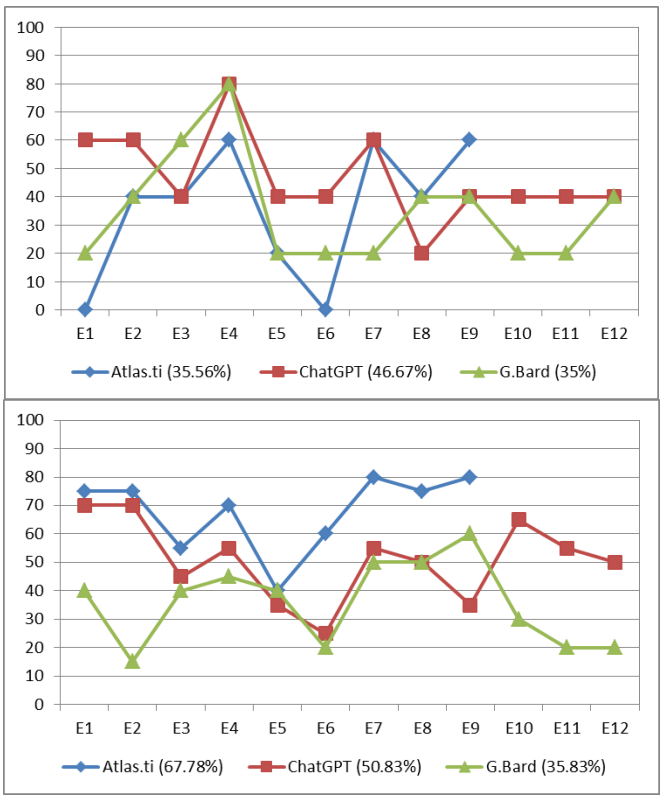

Source: The author.

Atlas.ti exhibits insufficient comprehensiveness in its coding, which is closely related to the low level of accuracy of the software in identifying the covered topics as LQOs. However, the opposite occurs with subcodes, where this software covers a large part of the relevant subthemes, which can be explained by the high number of subcodes it offers. The downside of this is that many of them are irrelevant and should not have been included (illustrated in Figure 8). Among the two conversational bots, ChatGPT performed better on this criterion (around 50%) than G. Bard (35%), both in codes and subcodes.

**FIGURE 7.** *Average percentage of the criteria Thematic Coherence (top) and Redundancy (bottom). The legend shows the average percentage for the set of interviews.*

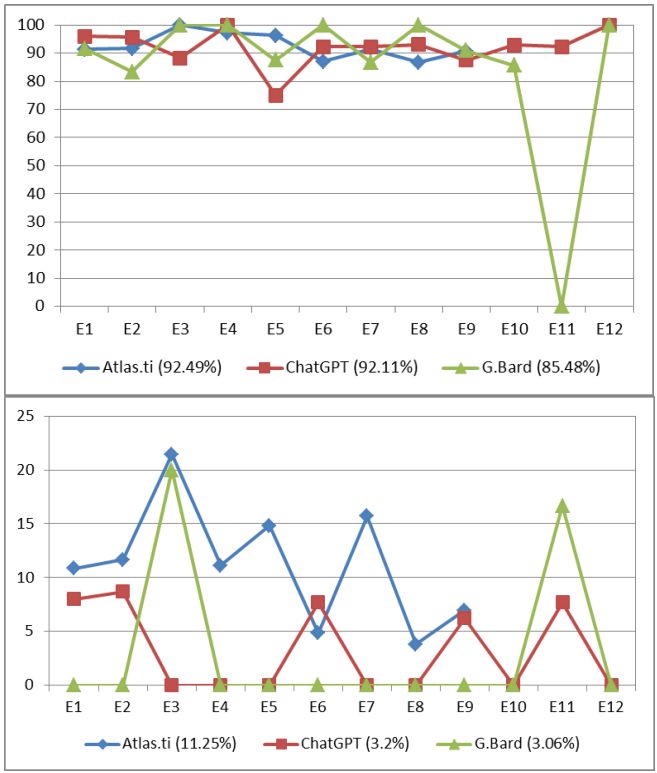

Source: The author.

Thematic Coherence was high in all cases (Figure 7, above), except in Interview 11, where Google Bard provided codes and subcodes separately (making it impossible to discern the corresponding relationships). The most optimistic interpretation of the high overall coherence percentage suggests that the AIs accurately comprehend the input text content and establish the correct relationships between codes and subcodes.

Redundancy analysis yields generally low percentages for this criterion, although higher for Atlas.ti (Figure 7, below). The AI incorporated into this software often provides an exact or nearly exact match (e.g.: "environmental conservation" and "environmental preservation"), which affects all possible combinations (code-code, code-subcode or subcode-subcode). Unfortunately, this is not adequately processed, leading to the misidentification of different codes or subcodes that are, in fact, the same. The absence of a higher average Redundancy in this software can be attributed solely to the great number of codes and subcodes it offers (e.g.: 10 redundancies out of 70 total items may represent a relatively low percentage, but it remains a very high number for a single interview) (Figure 8).

**FIGURE 8.** *Automatic coding in Atlas.ti for interviews (from left to right) 2, 3, 7 and 8. The high number of subcodes and the presence of redundancy can be observed.*

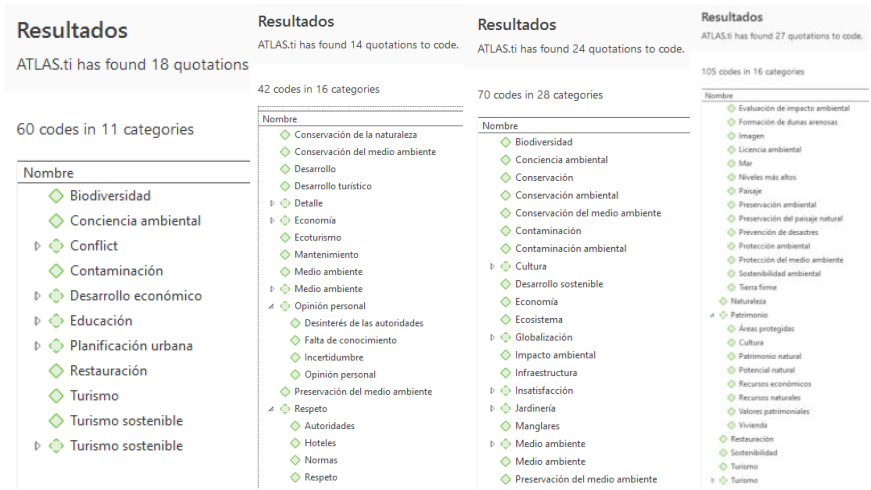

Source: The author based on the software Atlas.ti 2023.

Clarity and Detail are also two interrelated criteria that should show balance in the responses (neither too simple, nor too long and confusing). In the analysed interviews, ChatGPT demonstrates a high level of Clarity (rating around 4, on a scale of 1 to 5). It consistently deliv-ers well-structured answers with a limited number of codes and sucodes. At the same time, it tends to give brief but sufficient explanations in each case. Google Bard exhibits a similar type of response to ChatGPT, but it is highly variable in format, sometimes confusing and overly lengthy. Consequently, it can be considered of average clarity. Atlas.ti represents the opposite extreme, usually providing codes and subcodes consisting of one or two words (with a few exceptions). This gives clarity to the answers, but the large number of subcodes offered by this AI and the remarkable level of redundancy between them adds confusion and hinders the interpretation, so it does not pass a medium level of Clarity.

Regarding the level of Detail, the programming integrated into Google Bard is to be valued positively, as it usually offers a rich explanation of the codes and subcodes, as well as its own analysis, conclusions, complementary information taken from the web, etc. This can be useful for those new to qualitative content analysis, as long as such analyses and conclusions are not taken as good without first confronting them with the literature and after a substantiated assessment. Notably, it has been found that the bibliographical sources mentioned by Bard do not exist in a large number of cases.

Moving to the other two AI systems, ChatGPT is the most balanced, with brief but sufficient explanations and details of the codes. It is worth remembering that chatbots can always be prompted to expand the answers or a part of them, so simplicity is preferable in principle. In contrast, Atlas.ti lacks detail or explanation, as its codes typically consist of one or two words. Unlike chatbots, direct information requests about codes are not possible in Atlas.ti. The software as a whole does offer valuable tools for in-depth analysis and work with codes (such as code reports, word clouds, and concept mining), but these are functions outside the AI section, once the codes are already available. Summarily, it could be said that in Bard the level of detail is very high, in ChatGPT it is medium-high, and in Atlas.ti it is very low.

In terms of Regularity, during the tests comparing AI responses across the different interviews, Atlas.ti was the most regular, qualifying as a medium-high level. It is followed in regularity by ChatGPT, which offered very similar answers in all interviews (medium level). Google Bard demonstrated the most irregular pattern (medium-low level), presenting diverse response formats in different interviews: various types of classification and hierarchies of codes and subcodes, different extension or degree of detail, etc. This irregularity should not be confused with the functionality of offering several versions of the same answer, which differentiates it from the other two AIs and can be very useful.

On the Regularity across different requests for the same interview, Atlas.ti could be classified as medium-high, demonstrating slightly more consistency than the other two AIs by providing a similar type and number of codes and subcodes. Conversely, the regularity in the two chatbots could be characterized as medium, with greater variability between the responses offered to different requests with the same input data.

This is a negative point in the context of scientific research, where it is crucial to be able to replicate experiments and obtain the same results under equal conditions. In ChatGPT and Google Bard it might be understandable that this happens, since they are tools designed to hold a conversation (language models) and, as happens between people, the contents of the same conversation occurring at two different times can vary. However, in Atlas.ti this is a less acceptable shortcoming, since it is a very specialised software and almost exclusively oriented towards research. In any case, along with other shortcomings already mentioned, the irregularity is an additional reason why the codings currently generated by AIs should not be taken for granted without careful scrutiny from an expert's perspective.

A final remark should be made with respect to the results obtained between the different interviews. It has been observed that the quality of the input data greatly affects the responses provided by the AIs. For instance, in Interview 4 there were elevated levels of Accuracy, Completeness and Thematic Coherence (positive attributes) coupled with low levels of Redundancy (negative attribute). This is mainly due to the fact that the interviewee's answers were clearer, more organised and closer to what was being asked. In contrast, when dealing with transcripts where topics and subtopics are mixed, or appear several times at different points in the response, AIs do not perform as well. This underscores the need for substantial improvement to reach human proficiency in interpreting complex texts.

4.2. SUMMARY OF ADVANTAGES AND DISADVANTAGES OF THE AIS

Table 1 provides a summary of the main advantages and disadvantages of the AI tools examined.

**TABLE 1**. *Main pros and cons of the AI tools examined.*

| | Advantages | Disadvantages |
|---|---|---|
| Atlas.ti | 1) High Thematic Coherence.<br>2) Medium-high Regularity of responses.<br>3) Medium level of Clarity.<br>4) Ease of use of automatic coding without prior knowledge.<br>5) Possibility of coding more than one document at a time.<br>6) Multiple options for qualitative analysis in addition to AI coding (Co-Occurrence, Code reports, Networks...). | 1) Low Accuracy and Comprehensiveness.<br>2) Excessive Redundancy<br>3) Very low level of Detail.<br>4) Excessive number of categories and codes (codes and subcodes in the other AIs).<br>5) The option to choose the number of categories does not always work (in the free version).<br>6) It does not learn after successive attempts.<br>7) Allows a limited number of documents to be encoded and a limited amount of text (trial version). |
| ChatGPT | 1) High Thematic Coherence.<br>2) Medium Accuracy and Comprehensiveness.<br>3) Low Redundancy.<br>4) Clarity and conciseness of answers.<br>5) Correct level of Detail (adequate length).<br>7) Appropriate suggestions.<br>8) It learns as more queries are made. | 1) Requires basic knowledge of how to write prompts to get the right results.<br>2) Medium regularity in the content of the answers.<br>3) In the free version, a limited amount of text can be processed. |
| Google Bard | 1) Medium Accuracy and Comprehensiveness.<br>2) High Thematic Coherence.<br>3) Low Redundancy.<br>4) Medium Clarity of response.<br>5) Very high level of Detail.<br>6) Several versions of response.<br>7) Offers own analysis, conclusions, etc.<br>8) The length and tone of the response can be modified: shorter, longer, simpler, more formal...<br>9) It learns as more queries are made.<br>10) Compares the results with other research. | 1) Requires basic knowledge of how to write prompts to get the right results.<br>2) Medium-low Regularity in the content of the answers.<br>3) The multiplicity of answer options and perspectives could confuse the user.<br>3) In the free version, a limited amount of text can be processed. |

Source: The author.

## 5. CONCLUSIONS

In this study, we conducted a comparative analysis of the automated coding provided by three Artificial Intelligence functionalities (Atlas.ti, ChatGPT, and Google Bard) in relation to the manual coding of 12 research interviews focused on Landscape Quality Objectives for a small island in the north of Cuba (Cayo Santa María).

Following the established comparison criteria, it can be observed that Atlas.ti yielded relatively low Accuracy results, which were comparatively more acceptable in the two chatbots. Secondly, all three AIs generally demonstrate a medium level of Comprehensiveness and a high level of Thematic Coherence. Although Redundancy shows overall low percentages, it remains a flaw that can hinder the coding process and should be corrected, particularly in the case of Atlas.ti. Regarding Clarity and Detail criteria, ChatGPT has proven to be the most balanced. In these aspects, Atlas.ti appears overly simplistic while Google Bard tends to be more wordy than necessary. Finally, the Regularity of AI responses can also be improved, especially for academic and research purposes, although here Atlas.ti comes out a little better than chatbots.

Based on the conducted analysis, the usefulness of AI has been confirmed as support for the coding of research interviews, an often long and tedious process. However, even taking into consideration that the three tools have been evaluated in their free version (in development or beta phase[2]), the coding they provide exhibits numerous errors and limitations. Therefore, these AIs should be used with caution and bearing in mind that the results they yield may not be valid in many cases.

---

[2] At the time of completion of this study (December 2023), Google has just taken a new step in the AI career, launching its Gemini proyect, with which this company intends to surpass OpenAI's ChatGPT-4. For the moment, Gemini is only available in English, integrated into Google Bard. But it is expected that the update rate of these tools will be very high, so they will require further evaluation to verify their progress both in what concerns this study (automatic interview coding) and for any other application.

In this sense, the combination of several AIs may be more beneficial than choosing one in particular, especially if they have different characteristics that complement each other. As this is such a recently developed field, rapid evolution is expected to bring the necessary improvements to these tools.

In summary, today the automatic coding of AIs can be considered useful as a guide towards a subsequent in-depth and meticulous analysis of the information by the researcher. At that point, it is the expert's judgement that should prevail, with the final decisions being made on the basis of the experience and knowledge acquired. Moreover, adherence to research ethics must consistently steer the course of the work.

# 6. REFERENCES

Burgui-Burgui, M., Ibarra Benlloch, P. & Echeverría Arnedo, M. T. (2017). Determinación de objetivos de calidad del paisaje mediante participación de actores sociales en Cayo Santa María (Villa Clara, Cuba). In: E. Chuvieco, M. Burgui-Burgui (Eds.), *Valores y Compromisos en la Conservación Ambiental, Actas del I Congreso Español de Ecoética*. Cáetedra de Ética Ambiental. Universidad de Alcalá. pp. 96–100.

Burgui-Burgui, M., Ibarra Benlloch, P. & Echeverría Arnedo, M. T. (2018). Evolución de la calidad del paisaje a partir del desarrollo turístico en Cayo Santa María (Villa Clara, Cuba). *Boletín de la Asociación de Geógrafos Españoles*, 78, 444-473. http://dx.doi.org/10.21138/bage.2720

Chandra, Y., Shang, L., Chandra, Y. & Shang, L. (2019). Computer-assisted qualitative research: An overview. In: Y. Chandra & L. Shang (Eds.), *Qualitative Research Using R: A Systematic Approach*. Springer. https://doi.org/10.1007/978-981-13-3170-1_2

Conde, F. (2010). *Análisis sociológico del sistema de discursos*. *Cuadernos Metodológicos, nº 43*. Centro de Investigaciones Sociológicas.

Council of Europe (2000). *The European Landscape Convention*. European Union. https://bit.ly/3T7dLz0

Google (s.f.). *Bard FAQ*. https://bard.google.com/faq

Graaf, R. & Vossen, R. (2013). Bits versus brains in content analysis. Comparing the advantages and disadvantages of manual and automated methods for content analysis. *Communications*, 38, 433-443. https://doi.org/10.1515/commun-2013-0025.

Hacking, C., Verbeek, H., Hamers, J.P.H. & Aarts, S. (2023) Comparing text mining and manual coding methods: Analysing interview data on quality of care in long-term care for older adults. *PLoS ONE*, 18(11): e0292578. https://doi.org/10.1371/journal.pone.0292578

Lopezosa, C. & Codina, L. (2023). *ChatGPT y software CAQDAS para el análisis cualitativo de entrevistas: pasos para combinar la inteligencia artificial de OpenAI con ATLAS.ti, Nvivo y MAXQDA*. Universidad Pompeu Fabra Serie Editorial DigiDoc. PCUV04/2023.

Lopezosa, C., Codina, L. & Boté-Vericad, J. J. (2023). *Testeando ATLAS.ti con OpenAI: hacia un nuevo paradigma para el análisis cualitativo de entrevistas con Inteligencia artificial*. Universidad Pompeu Fabra Serie Editorial DigiDoc. PCUV05/2023.

Medina, J. E. & Hernández, L. (2007). Segmentación por tópicos en documentos de múltiples párrafos. *Revista Acimed*, 15 (6).

OpenAI (2022). *Introducing ChatGPT*. https://openai.com/blog/chatgpt.

Singh, S. K., Kumar, S. & Mehra, P. S. (2023). *Chat GPT & Google Bard AI: A Review*. International Conference on IoT, Communication and Automation Technology (ICICAT) (pp. 1-6). IEEE.

Skjott Linneberg, M. & Korsgaard, S. (2019), Coding qualitative data: a synthesis guiding the novice. *Qualitative Research Journal*, 19 (3), 259-270. https://doi.org/10.1108/QRJ-12-2018-0012

Soliveres, M. A., Anunziata, S. M. & Macías, A. (2007). La comprensión de la idea principal de textos de Ciencias Naturales. Una experiencia con directivos y docentes de EGB2. *Revista Electrónica de Enseñanza de las Ciencias*, 6 (3), 577-586.

White, J., Fu, Q., Hays, S., Sandborn, M., Olea, C., Gilbert, H., Elnashar, A., Spencer-Smith, J. & Schmidt, D. C. (2023). A prompt pattern catalog to enhance prompt engineering with chatgpt. *arXiv preprint*. https://doi.org/10.48550/arXiv.2302.11382

Zhang, H., Wu, C., Xie, J., Lyu, Y., Cai, J., & Carroll, J. M. (2023). Redefining qualitative analysis in the AI era: Utilizing ChatGPT for efficient thematic analysis. *arXiv preprint*. https://doi.org/10.48550/arXiv.2309.10771